\documentclass{article}
\usepackage{arxiv}
\usepackage[utf8]{inputenc} 
\usepackage[T1]{fontenc}
\usepackage{graphicx}
\usepackage[colorlinks=true, linkcolor=black, urlcolor=black, citecolor=black]{hyperref}
\usepackage{color}

\usepackage{booktabs}       
\usepackage{subcaption}
\usepackage{caption}
\usepackage{amsfonts}
\usepackage{csquotes}
\usepackage{todonotes}  
\usepackage{verbatim}
\usepackage[nolist]{acronym}
\usepackage{fontawesome5}

\usepackage
[backend=biber,
 maxbibnames=6,
 minbibnames=3,
 style=authoryear
]{biblatex}
\usepackage{doi}

\usepackage{xurl}
\usepackage{hyperref}
\hypersetup{breaklinks=true} 

\usepackage[linesnumbered,ruled,longend]{algorithm2e}
\usepackage{amsmath}
\usepackage{amsfonts}
\usepackage{rotating}
\usepackage{marvosym}

\newcommand{\orcidID}[1]{\unskip\href{https://orcid.org/#1}{\textsuperscript{\includegraphics[scale=0.06]{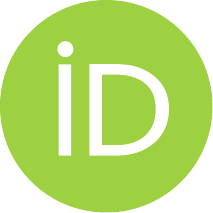}}}}

\begin{document}
\title{Demonstrating Data-to-Knowledge Pipelines for Connecting Production Sites in the World Wide Lab}
\renewcommand{\shorttitle}{Data-to-Knowledge Pipelines for Connecting Production Sites in the World Wide Lab}
%

\author{
    ~\href{https://www.llt.rwth-aachen.de/cms/llt/der-lehrstuhl/mitarbeiter/~bikmgq/leon-gorissen/?allou=1}{Leon Gorißen\textsuperscript{\Letter}}\orcidID{0000-0002-6696-465X}$^{1}$,\and
    Jan-Niklas Schneider\orcidID{0009-0003-5781-6270}$^{1}$,\and
    Mohamed Behery\orcidID{0000-0003-1331-9419}$^{2}$,\and
    Philipp Brauner\orcidID{0000-0003-2837-5181}$^{3}$,\and
    Moritz Lennartz\orcidID{0009-0001-9072-2511}$^{4}$,\and
    David Kötter\orcidID{0009-0002-6552-190X}$^{5}$,\and
    Thomas Kaster\orcidID{0000-0003-0381-2277}$^{1}$,\and
    Oliver Petrovic\orcidID{0000-0002-4861-1332}$^{5}$,\and
    Christian Hinke\orcidID{0009-0006-7382-1987}$^{1}$,\and
    Thomas Gries\orcidID{0000-0002-2480-8333}$^{4}$,\and
    Gerhard Lakemeyer\orcidID{0000-0002-7363-7593}$^{2}$,\and
    Martina Ziefle\orcidID{0000-0002-6105-4729}$^{6}$,\and
    Christian Brecher\orcidID{0000-0002-8049-3364}$^{5}$,\and
    Constantin Häfner\orcidID{0000-0002-7965-1708}$^{1,7}$\\
    \\
    $^{1}$Chair for Laser Technology, RWTH Aachen University, 52074 Aachen, Germany\\
    $^{2}$Knowledge Based Systems Group, RWTH Aachen University, \\
    $^{3}$Communication Science, RWTH Aachen University, \\
    $^{4}$Chair of Textile Technology, RWTH Aachen University, \\
    $^{5}$Laboratory for Machine Tools and Production Engineering, RWTH Aachen University, \\
    $^{6}$Human Computer Interaction Center, RWTH Aachen University, \\
    $^{7}$Fraunhofer Institute for Laser Technology, 52074 Aachen, Germany\\
    \\
    \texttt{leon.gorissen@llt.rwth-aachen.de}
}

\maketitle              


\begin{acronym}
    \acro{erp}[ERP]{Enterprise Resource Planning}
    \acro{dss}[DSS]{Decision Support System}
    \acroplural{dss}[DSSs]{Decision Support Systems}
    \acro{dag}{directed acyclic graph}
    \acroplural{dag}{directed acyclic graphs}  
    \acro{iop}[IoP]{Internet of Production}
    \acro{cnn}[CNN]{Convolutional Neural Network}
    \acroplural{cnn}[CNNs]{Convolutional Neural Networks}
    \acro{sme}[SME]{small and me\-di\-um-sized enterprise}
    \acroplural{sme}[SMEs]{small and me\-di\-um-sized enterprises}
    \acro{ds}[DS]{Digital Shadow} 
    \acroplural{ds}[DSs]{Digital Shadows} 
    \acro{dt}[DT]{Digital Twin} 
    \acroplural{dt}[DTs]{Digital Twins} 
    \acro{ml}[ML]{Machine Learning} 
    \acro{bt}[BT]{Behavior Tree}
    \acro{bts}[BTs]{Behavior Trees}
    \acro{htns}[HTNs]{Hirarichal Task Networks}
    \acro{htn}[HTN]{Hirarichal Task Network}
    \acro{fsm}[FSM]{Finite State Machine}
    \acro{fsms}[FSMs]{Finite State Machines}
    \acro{plc}[PLC]{Product Life Cycle}
    \acro{dcc}[DCC]{Digital Capability Center}
    \acro{bol}[BOL]{Beginning of Life}
    \acro{ai}[AI]{Artificial Intelligence}
    \acro{xai}[XAI]{eXplainable Artificial Intelligence}
    \acro{hri}[HRI]{Human Robot Interaction}
    \acro{hrc}[HRC]{Human Robot Collaboration}
    \acro{iot}[IoT]{Internet of Things}
    \acro{iiot}[IIoT]{Industrial Internet of Things}
    \acro{www}[WWW]{World Wide Web}
    \acro{wwl}[WWL]{World Wide Lab}
    \acro{frp}[FRP]{Fiber Reinforced Plastics}
    \acro{sus}[SUS]{System Usability Scale}
    \acro{tfd}[TFD]{Temporal Fast Downward}
    \acro{pddl}[PDDL]{Planning Domain Definition Language}
    \acro{lmp}[LMP]{Laser Material Processing}
    \acro{lstm}[LSTM]{Long-Short-Term-Memory}
    \acro{ir}[IR]{Industrial Robot}
    \acroplural{ir}[IRs]{Industrial Robots}
    \acro{tcp}[TCP]{Tool Center Point}
    \acro{el}[EL]{Euler Lagrange}
    \acro{rnn}[RNN]{Recurrent Neural Network}
    \acro{rel}[RL]{Reinforcement Learning}
    \acro{am}[AM]{Additive Manufacturing}
    \acro{lpbf}[LPBF]{Laser Powder Bed Fusion}
    \acro{ae}[AE]{Auto Encoder}
    \acro{dl}[DL]{Deep Learning}
    \acro{cpps}[CPPS]{Cyber-Physical Production System}
    \acro{mtl}[MTL]{Metric Temporal Logic}
    \acro{ocel}[OCELs]{Object-Centric Event Log}
    \acroplural{ocel}[OCELs]{Object-Centric Event Logs}
    \acro{gan}[GAN]{Generative Adversarial Network}
    \acroplural{gan}[GANs]{Generative Adversarial Networks}
    \acro{ann}[ANN]{Artificial Neural Network}
    \acroplural{ann}[ANNs]{Artificial Neural Networks}
    \acro{ot}[OT]{Optical Tomography}
    \acro{kpi}[KPI]{Key Performance Indicator}
    \acroplural{kpi}[KPIs]{Key Performance Indicators}
    \acro{ocpc}[OCPC]{Object-Centric Process Cube}
    \acroplural{ocpc}[OCPCs]{Object-Centric Process Cubes}
    \acro{nlp}[NLP]{Natural Language Processing}
    \acro{dag}[DAG]{Directed Acyclic Graph}
    \acroplural{dag}[DAGs]{Directed Acyclic Graphs}
    \acro{fair}[FAIR]{Findable, Accessible, Interoperable, Reusable}
    \acro{fdo}[FDO]{FAIR Digital Object}
    \acro{uuid}[UUID]{Universally Unique Identifier}
    \acro{pinn}[PINN]{Physics-Informed Neural Network}
    \acroplural{pinn}[PINNs]{Physics-Informed Neural Networks}
    \acro{mae}[MAE]{mean absolute error}
    \acro{d2k}[D2K]{Data-to-Knowledge}
    \acro{k2d}[K2D]{Knowledge-to-Data}
    \acro{mes}[MES]{Manufacturing Execution System}
    \acro{llt}[LLT]{Chair for Laser Technology}
    \acro{ita}[ITA]{Institute for Textile Technology}
    \acro{wzl}[WZL]{Chair for Machine Tools}
\end{acronym}

\begin{abstract}
The digital transformation of production requires new methods of data integration and storage, as well as decision making and support systems that work vertically and horizontally throughout the development, production, and use cycle.
In this paper, we propose \acl{d2k} (and \acl{k2d}) pipelines for production as a universal concept building on a network of \aclp{ds} (a concept augmenting \aclp{dt}).
We show a proof of concept that builds on and bridges existing infrastructure to 1) capture and semantically annotates trajectory data from multiple similar but independent robots in different organisations and use cases in a data lakehouse and 2) an independent process that dynamically queries matching data for training an inverse dynamics foundation model for robotic control.
The article discusses the challenges and benefits of this approach and how \acl{d2k} pipelines contribute efficiency gains and industrial scalability in a \acl{wwl} as a research outlook.

\end{abstract}

\section{Introduction}\label{sec:introduction}

Data is crucial for the digital transformation of production as it enables real-time insights, (automatic) process optimization, and the integration of \ac{ai}, \ac{iot} and other advanced technologies, leading to increased efficiency, innovation, and more sustainable production~\autocite{Bruner2013,Kagermann2015}.
However, as production environments become increasingly interconnected, we face challenges in data integration in and between companies, data quality management, interoperability, and security across diverse, legacy, and often siloed systems~\parencite{Lenzerini2002, Omrany2023, Kadadi2014}.
Addressing these challenges requires conceptual frameworks that can bridge organizational boundaries and enable collaborative production ecosystems.

This paper introduces \ac{d2k} and \ac{k2d} pipelines for production.
These pipelines facilitate the bidirectional transformation of raw data from machine sensors for either fully automated systems or actionable insights for decision makers and vice versa.
By leveraging interconnected Digital Shadows (DSs)---modular, context-dependent representations of production entities---these pipelines provide the foundation for dynamic, adaptive, and scalable production systems.

We demonstrate that the pipelines enable the seamless capture, semantic annotation, and integration of trajectory data from robots operating in diverse organizations.
Using a shared research data repository, they support the training of foundational models for robotic control, which can then be fine-tuned for specific use cases.
This approach reduces data redundancy and enhances model robustness and scalability, aligning with the principles of FAIR (Findable, Accessible, Interoperable, Reusable) data and provides a pathway for organizations to exploit their data in the increasingly complex production landscape.


This article is structured as follows.
Section~\ref{sec:relatedwork} reviews the related work on data in production domains.
Section~\ref{sec:integrated_definitions_and_architecture} describes the architecture of the \ac{d2k} and \ac{k2d} pipelines.
Section~\ref{sec:realization} present a use-cases demonstrating the practical benefits of these pipelines. 
Section~\ref{section:discussion} discusses the findings, challenges, and broader implications of our approach.
Finally, Section~\ref{section:conclusion} concludes with an outlook on future research and the potential for these pipelines to advance the vision of the \ac{wwl}.

\section{Background and Related Work}\label{sec:relatedwork}

The industrial revolutions have advanced manufacturing from steam-powered machinery and mass production to digital automation and data-driven decision making. 
Industry 4.0 aims at integrating legacy systems, disparate systems, and data silos into cohesive frameworks to utilize data effectively \parencite{Lu2017,Kagermann2015,Zhong2017}.

\emph{Data architectures} evolved to address data challenges \parencite{Nargesian2019,Nambiar2022}. \emph{Data warehouses} centralize structured data for analytics \parencite{Inmon2002}, yet are task-specific, as seen with \ac{erp} systems, which lack broader integration with tools like \ac{mes}. \emph{Data lakes} store raw, multi-format data but pose querying complexity \parencite{dixon2010}. \emph{Data lakehouses} combine warehouse and lake capabilities \parencite{Harby2022}.
More recently, \emph{data meshes} have been proposed that enable decentralized, domain-specific data ownership \parencite{Goedegebuure2024} and \emph{data fabric} that integrates distributed data sources, enhancing governance \parencite{IBMDataFabric}. Table \ref{tab:comparison} compares these architectures.
\begin{sidewaystable} 
\caption{Comparison of Data Architectures}
\centering
\begin{tabular}{@{}p{3cm}p{3.2cm}p{3.2cm}p{3.2cm}p{3.2cm}p{3.2cm}@{}}
\toprule
\textbf{Aspect} & \textbf{Data Warehouse} & \textbf{Data Lake} & \textbf{Data Lakehouse} & \textbf{Data Mesh} & \textbf{Data Fabric} \\ \midrule

\textbf{Data Storage} 
& Structured, transformed data\newline\parencite{Vuka2018, Souissi2016, Ghosh2017} 
& Raw, unstructured data\newline\parencite{Belov2021, Sousa2023, Liu2020} 
& Combines structured and unstructured data\newline\parencite{Orescanin2021, Harby2025} 
& Decentralized, domain-oriented data\newline\parencite{Dolhopolov2024, Azeroual2023, Machado2021} 
& Distributed, integrates various sources\newline\parencite{Rajakumari2023, Kuftinova2023} \\ \midrule

\textbf{Data Processing} 
& ETL (Extract, Transform, Load)\newline\parencite{Vuka2018, Souissi2016} 
& Post-storage processing\newline\parencite{Sousa2023, Liu2020} 
& Post-storage transformation and analytics\newline\parencite{Orescanin2021, Harby2025} 
& Federated computational governance\newline\parencite{Dolhopolov2024, Azeroual2023} 
& AI-enabled integration and processing\newline\parencite{Debellis2023, Ciaccia2023} \\ \midrule

\textbf{Governance} 
& Centralized, schema-based\newline\parencite{Vuka2018, Souissi2016} 
& Minimal governance, risk of data swamps\newline\parencite{Sousa2023, Liu2020} 
& Enhanced governance combining DW and DL features\newline\parencite{Orescanin2021, Harby2025} 
& Decentralized, domain-specific governance\newline\parencite{Dolhopolov2024, Machado2021} 
& Centralized governance with distributed architecture\newline\parencite{Rajakumari2023, Kuftinova2023} \\ \midrule

\textbf{Use Cases} 
& Business intelligence, historical data analysis\newline\parencite{Vuka2018, Souissi2016} 
& Big data analytics, machine learning\newline\parencite{Sousa2023, Liu2020} 
& Unified analytics, real-time data processing\newline\parencite{Orescanin2021, Harby2025} 
& Data sharing and reuse, scalable data management\newline\parencite{Dolhopolov2024, Azeroual2023} 
& Seamless data access, compliance, and integration across platforms\newline\parencite{Debellis2023, Ciaccia2023} \\ \bottomrule

\end{tabular}
\label{tab:comparison}
\end{sidewaystable}

\subsection{Internet of Production and the World Wide Lab}\label{section:iopwwl}

The \ac{iop} and \ac{wwl} expand internal organizational data frameworks to encourage collaboration across supply chains and research labs \parencite{schuh2019internet, BeheryWWL2023}.
The \ac{iop} links various production entities from design to logistics, forming a network of industrial participants that integrates sensors, machines, and planning tools across supply chains.
Similarly, the \ac{wwl}, inspired by the \ac{www}, aims to provide a global platform for shared innovation, enabling cross-com\-pa\-ny, hierarchical, and horizontal data integration \parencite{BeheryWWL2023}.
It aims to promote information exchange, reduce redundant experimentation, and enhance the quality of data-driven models by incorporating empirical data from diverse manufacturing environments.

Rather than being specific implementations, the \ac{iop} and \ac{wwl} are conceptual frameworks adaptable to various data layer technologies. For instance, \ac{iop} initiatives  like \textit{ProducTron} focus on distributed, modular production, emphasizing flexibility within companies \parencite{Pallasch2018ProducTron}, while \textit{FactDAG} improves data interoperability through a conceptual data layer model based on provenance-based directed acyclic graphs, supporting \ac{wwl} goals of transparency and collaboration \parencite{Gleim2020FactDAG}.
Blockchain-based systems, such as the \textit{Trustworthy Information Store}, ensure accountability and verifiability in dynamic collaborations \parencite{Pennekamp2021Blockchain}.

Complementary initiatives like \textit{Gaia-X}, \textit{Catena-X}, and \textit{Manufacturing-X} also advance to industrial data sharing by promoting secure, federated infrastructures \parencite{GaiaX2020, CatenaX2021, ManufacturingX2022}.
These efforts align with the \ac{wwl}'s goals of enhancing interoperability, data sovereignty, and cross-industry collaboration, albeit with specific foci, such as the automotive industry or adherence to  European standards.

\subsection{From Digital Twins to Digital Shadows}\label{section:digitalshadows}

A cornerstone of the \ac{wwl} is the use of \acp{ds}, defined as \enquote{task- and context-dependent, purpose-driven, aggregated, multi-perspective, and persistent datasets} \parencite[pp. 3]{Brauner2022} and are comparable to views in relational databases \parencite{Liebenberg2020, Date2019}.
Unlike \acp{dt}, which are often tied to a single representation, \acp{ds} allow multiple projections of the same entity, enabling modularity and scalability \parencite{Liebenberg2020}:
For instance, a production entity might have one \ac{ds} for high-precision simulations and another for real-time control \parencite{behery2023digital-cms}.

\acp{ds} support hierarchical and cross-company scalability by functioning at multiple levels, such as individual machines or generalized machine types \parencite{bauernhansl2018digital}.
Their design is informed by principles of separation of concerns \parencite{kulkarni2003separation} and microservices architecture \parencite{nadareishvili2016microservice}, enhancing maintainability and adaptability. For example, operators can modify tasks, adapt to individualized production, and manage product variants by updating specific \acp{ds}, reducing memory and processing demands while maintaining contextual relevance.
By forming networks of interconnected \acp{ds}, the \ac{wwl} enables tracking and analysis of data flows across \ac{d2k} and \ac{k2d} pipelines, enhancing transparency and decision-making interfaces.

Moreover, \textit{FactDAG} exemplifies how \acp{ds} can leverage directed acyclic graphs to maintain data provenance, ensuring transparency and alignment with FAIR data principles \parencite{Gleim2020FactDAG}.
This approach facilitates data-driven innovation while maintaining trust and accountability in collaborative production environments.

Thus, the \ac{wwl} and \acp{ds} enable industrial big data to become actionable through innovative, efficient, and modular \ac{d2k} and \ac{k2d} pipelines. These systems empower informed decisions by allowing users and processes to analyze data flows at any level, promoting transparency and contextual adaptability for modern manufacturing systems.

\section{Data, Knowledge, Autonomous Agents, and Transformative Pipelines}\label{sec:integrated_definitions_and_architecture}

\ac{d2k} and \ac{k2d} pipelines facilitate the seamless conversion of raw information into actionable insights and decision-making processes by transforming \emph{data} into \emph{knowledge} for \emph{autonomous agents}.

\emph{Data} refers to raw, context-dependent information gathered from sources such as sensors, machines, and humans. It can be in structured, semi-structured, or unstructured formats, including measurements, logs, and images. Data undergo transformations such as cleaning and semantic annotation to become actionable knowledge.

\emph{Knowledge} is actionable, semantically enriched information derived from data or inherent in autonomous agents. It aids decision-making, system optimization, and continuous improvement through analysis and interpretation. Through autonomous agents, knowledge can direct both process control and system-level decisions.

\emph{Autonomous agents} are entities---both human and artificial---that can perceive their environment, process data, make decisions, and adapt over time. These agents act as intermediaries between data and knowledge, enabling responsive and adaptive systems.

The definitions presented here align with foundational perspectives in the literature on knowledge management, such as those of \textcite{Alavi2001}, which emphasize data as raw, context-dependent input transformed into actionable knowledge through semantic enrichment and autonomous agents as critical mediators that perceive, process and adapt to enable decision-making and responsive systems.

\begin{figure}[htbp]
\centering
\includegraphics[width=\textwidth]{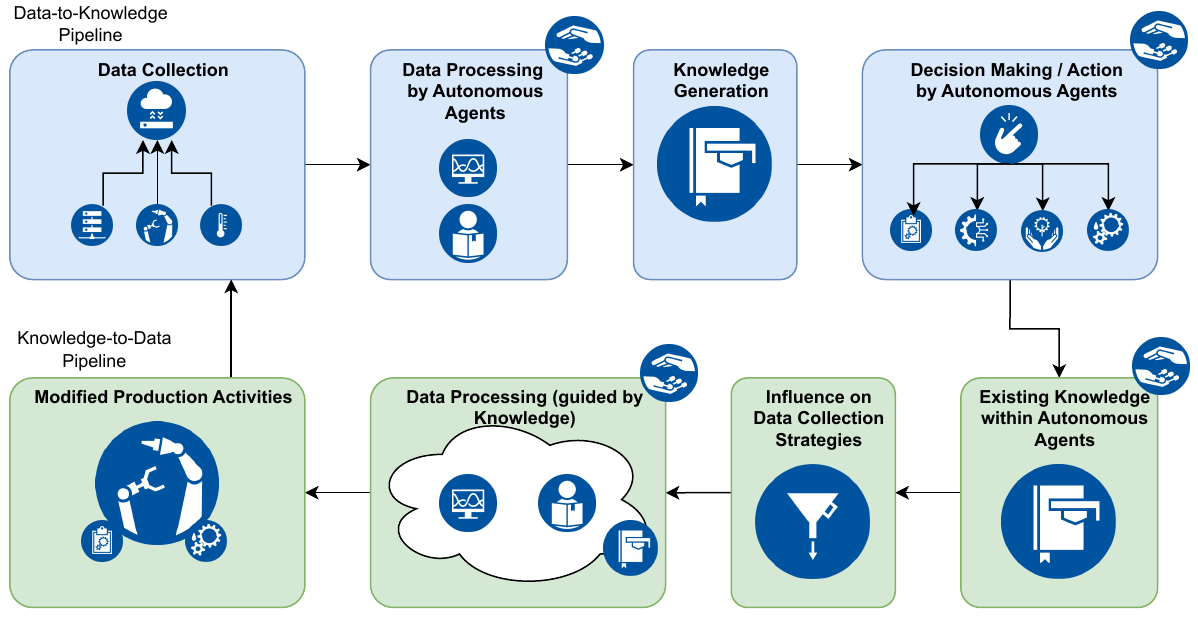}
\caption{Illustration of the steps within the proposed pipelines. Each pipeline itself is a directed acyclic graph, but Data-to-Knowledge Pipeline can be a start to create a Knowledge-to-Data Pipeline and vice versa.}
\label{fig:pipelineConcept}
\end{figure}

Figure \ref{fig:pipelineConcept} illustrates the interplay of the proposed pipelines, demonstrating how they facilitate integration from the shop-floor to decision-making. Using analytics and machine learning, these pipelines form adaptive networks that optimize operations, reduce costs, and drive innovation.

The \ac{d2k} pipeline converts raw data into knowledge through a series of steps: 
\begin{enumerate}
    \item \textbf{Data Collection} from various sources.
    \item \textbf{Data Processing} to transform through cleaning and annotation.
    \item \textbf{Knowledge Generation} using analytics and learning methods.
    \item \textbf{Action} application for process optimization and decision-making.
\end{enumerate}
Such application of knowledge may be a starting point to \ac{k2d} pipelines.
Conversely, the \ac{k2d} pipeline applies knowledge to guide data collection and system adjustments. Its steps are:
\begin{enumerate}
    \item \textbf{Existing Knowledge} used to inform data collection.
    \item \textbf{Influence on Data Collection Strategies}.
    \item \textbf{Data Processing} shaped by knowledge.
    \item \textbf{Modified Production Activities}.
\end{enumerate}
These \ac{k2d} pipelines may again form a starting point for a \ac{d2k} given they adapt the data source.

These pipelines facilitate seamless integration from shop-floor machinery to corporate decision-making, forming an overarching framework for capturing, analyzing, and applying data and knowledge to optimize operations and drive business outcomes.
For example, a \ac{ d2k} pipeline might enable autonomous control of a production robot by processing sensor values filtered through a \ac{ds}, or it could extend across multiple layers to facilitate decision-making on live production data at the organizational level.
These pipelines leverage \acp{ds},  predictive analytics, machine learning algorithms, and other techniques to extract actionable knowledge. This knowledge can be used for smarter control loops, supporting shop-floor workers, or guiding decision-makers in optimizing production strategies.

While simple \ac{d2k} pipelines may follow a linear structure, they can also form complex networks spanning \acp{dag}. Such architectures allow the integration of multiple data sources or systems into a unified decision-making process.
For instance, sensor data from different production units could be aggregated to optimize a control loop, or foundational aggregated data might be utilized by various stakeholders to derive insights specific to their needs.

When interconnected and automated, these processes establish a network of \acp{ds} that continuously capture, analyze, and apply data and knowledge. This network enables organizations to leverage real-time data and insights to enhance efficiency, reduce costs, and drive innovation. Moreover, the integration of \ac{d2k} and \ac{k2d} pipelines ensures a cyclical flow of improvement, where data informs knowledge generation, and knowledge, in turn, guides data collection and processing. These pipelines can span various organizational processes and supply chain nodes, enabling coordinated optimization efforts. They support multi-layered integration, with vertical integration linking shop-floor systems to executive decision-making, and horizontal integration facilitating collaboration across organizations. Autonomous systems and interconnected \acp{ds} provide real-time responsiveness, ensuring timely data processing and decision-making to enhance operational agility. By creating collaborative networks, these pipelines reduce redundancy, enrich models, and promote collective optimization. The dynamic interplay of \ac{d2k} and \ac{k2d} pipelines further encourages adaptive learning, innovation, and sustained system optimization, ensuring continuous improvement in interconnected environments.

This integrated framework supports collaboration, real-time responsiveness, and sustained improvement, facilitating modern production environments and innovation.
The following section illustrates the practical realization of these pipelines, showcasing their benefits through real-world examples and their implementation within the \textit{World Wide Lab}.

\section{Realisation of Data-To-Knowledge Pipelines}\label{sec:realization}

\begin{figure}[htbp]
\centering
\includegraphics[width=\textwidth]{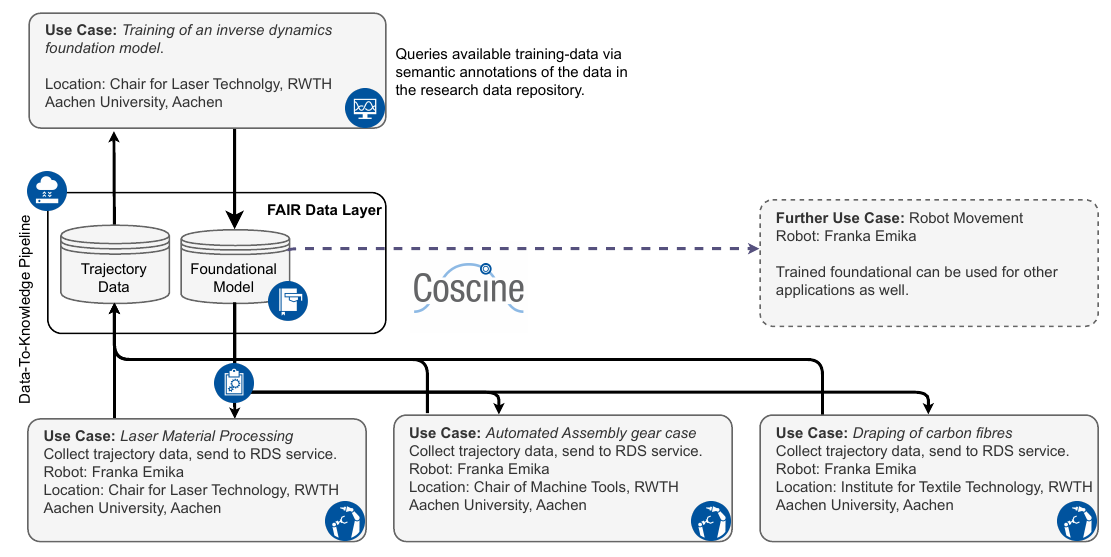}
\caption{In this network of \acl{d2k} pipelines trajectory data from multiple sources of different organizations are stored in a research data management repository and semantically annotated (lower use cases). The training instance (upper use case) queries available training data and provides a foundation model. Instance specific models are derived from the foundation model by the original use cases or third party use cases.}
\label{fig:pipelines}
\end{figure}

To demonstrate the benefits of \ac{d2k} pipelines, we integrate data from three Franka Emika robots operating at different organizations for a machine learning task.
Each robot performs specific tasks and collects trajectory data, which is used to train a foundation model on robot dynamics.
This model enables deriving instance-specific models, that can be used in torque-based control strategies. Figure~\ref{fig:pipelines} outlines the demonstrated \ac{d2k} pipeline within the \ac{wwl}. Code and data are available as part of the replication package under \url{https://s.fhg.de/gorissen-2025a}.

Instead of creating an additional data silo and future legacy system, we build on our university's centralized research data infrastructure (coscine~\parencite{coscine}) to manage data storage, findability, and accessibility.
This approach streamlines data handling, providing automatic management, broader datasets for training foundation models, and compliance with \acs{fair} principles \parencite{Bodenbenner2023}.
In the \ac{iop}, we advocate using \acsp{ds} for their task-specific data views over traditional \acsp{dt} \parencite{Brauner2022, Liebenberg2020}.
Consequently, we use existing infrastructure to integrate \acsp{ds} across different organisations. 


At the machine level, \acp{ds} represent specific robots through command and attained trajectory data, fine-tuned dynamics models, and virtual scenes. Machine-level \acp{ds} aggregate data across robot types, enabling knowledge transfer and scalability.

\subsection{Domain-specific Manufacturing Use Cases}

\begin{figure}[htbp]
    \centering
    \begin{subfigure}[t]{0.3\textwidth}
      \includegraphics[width=\textwidth]{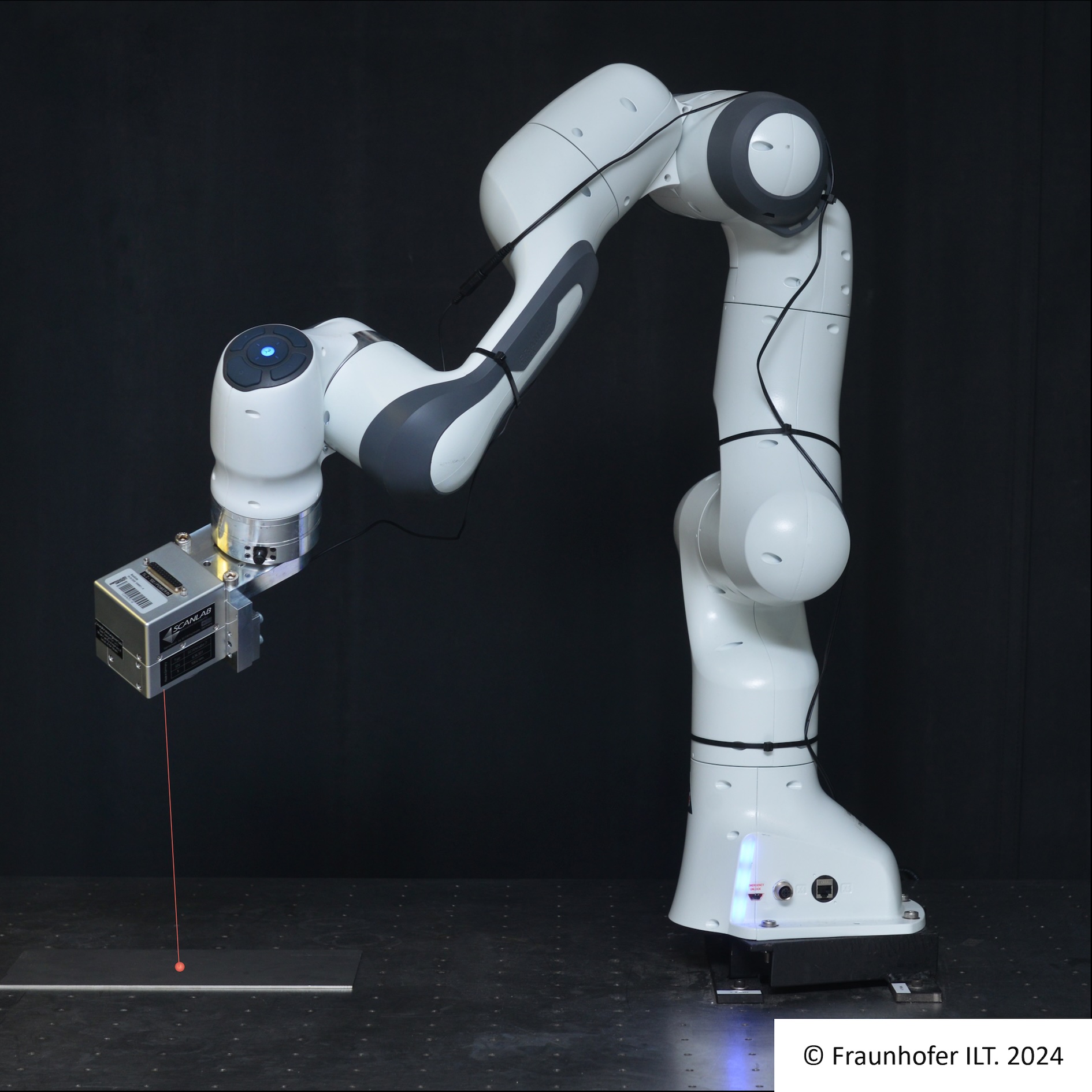}
      \caption{\acl{llt}: Franka Emika Robot used in laser material processing of steel. \label{fig:robots:llt}}
    \end{subfigure}\hfill
    \begin{subfigure}[t]{0.3\textwidth}
      \includegraphics[width=\textwidth]{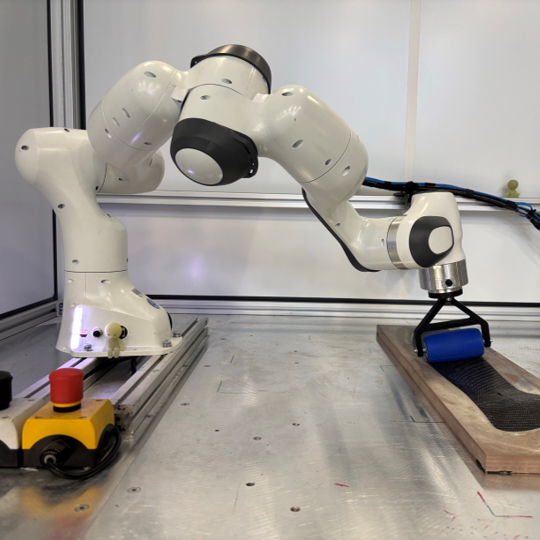}
      \caption{\acl{ita}: Franka Emika Robot used in automated draping of fiber composites. \label{fig:robots:ita}}
    \end{subfigure}\hfill
    \begin{subfigure}[t]{0.3\textwidth}
      \includegraphics[width=\textwidth]{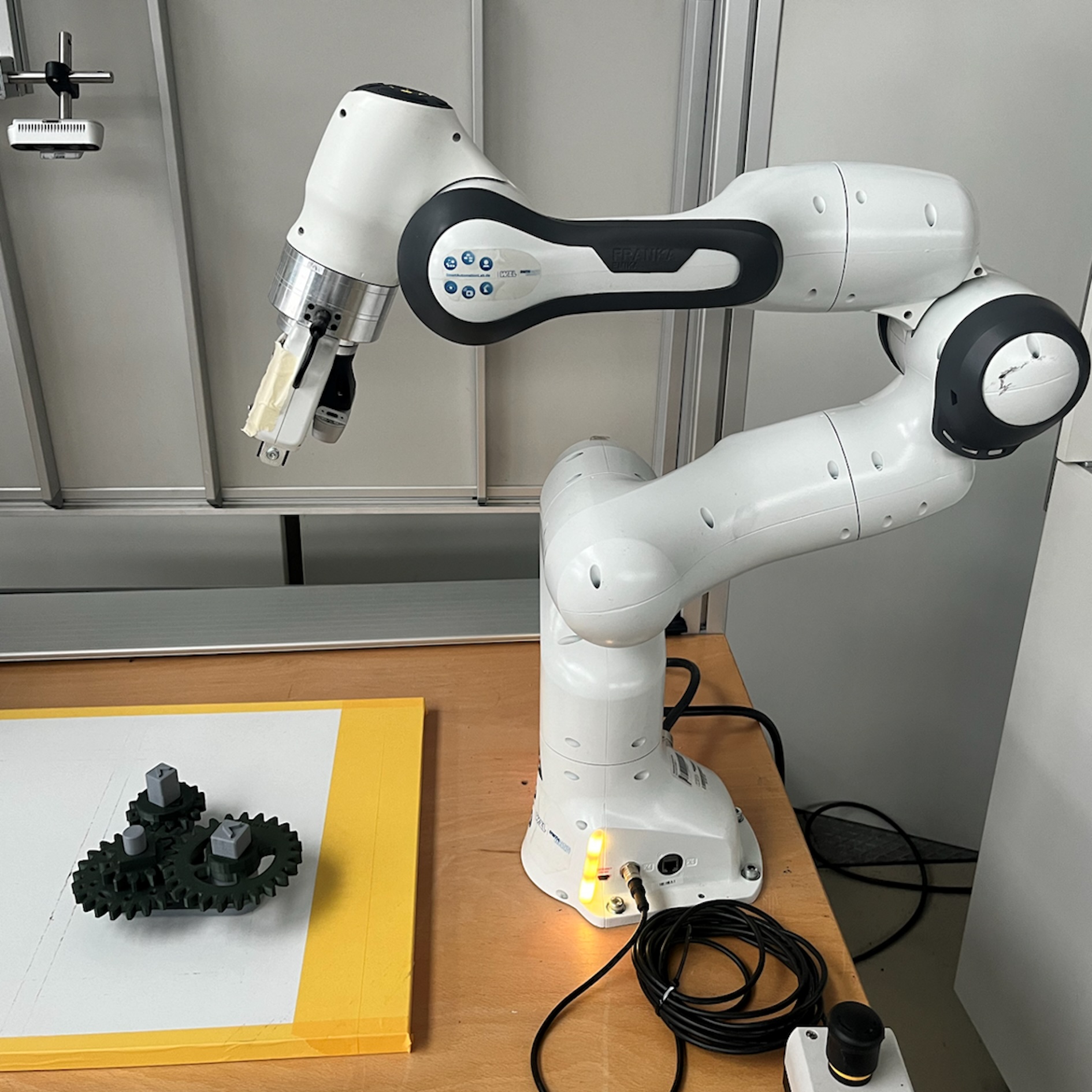}
      \caption{\acl{wzl}: Franka Emika Robot in collaborative gear assembly. \label{fig:robots:wzl}}
    \end{subfigure}
    \caption{Franka Emika Robots integrated into a \acl{d2k} pipeline to train a foundation model of robot dynamics.}
\end{figure}

\paragraph{Laser Material Processing Use Case}\label{paragraph:usecase:laser}
The \ac{llt} demonstrates laser engraving using a Franka Emika robot equipped with a fast beam steering device, addressing trajectory inaccuracies critical to geometric tolerances in laser-matter interactions \parencite{Schneider2024}. Figure~\ref{fig:robots:llt} shows the setup.

\paragraph{Textile Use Case}\label{section:usecase:textile}
The \ac{ita} automates fiber composite production using cobots for flexible manufacturing in \acp{sme}, preventing damage to delicate textiles \parencite{Dammers2024}. Figure~\ref{fig:robots:ita} depicts the cobot performing a draping process.

\paragraph{Assembly Use Case}\label{section:usecase:assembly}
The \ac{wzl} applies collaborative robotics for precise gear assembly. The robot solves peg-in-hole problems using inverse dynamics for accurate control \parencite{HRCtrends}. Figure~\ref{fig:robots:wzl} shows the setup.

\subsection{Inverse Dynamics Data-to-Knowledge Pipeline}

We demonstrate the concept of a \ac{d2k} pipeline by aggregating trajectory data from Franka Emika Robots used in diverse domains and tasks.
This data is leveraged to train a foundation model addressing the inverse dynamics problem---determining the joint torques or forces required to achieve a desired motion trajectory given the robot's configuration, velocities, and accelerations~\parencite{Siciliano.2016}---using a stacked \ac{lstm} neural network architecture, as previously proposed by Schneider et al. \parencite{Schneider2024}.
Foundation models, such as this one, serve as pre-trained starting points, supporting efficient fine-tuning for task-specific applications and reducing the need for extensive data collection in downstream tasks \parencite{bommasani2022opportunitiesrisksfoundationmodels}.

The foundation model generalizes across different instances of the Franka Emika Robot, enabling knowledge transfer between various use cases. Instance-specific fine-tuning further adapts the model for precise control in distinct production tasks. Unlike earlier approaches \parencite{Schneider2024, Kruzic2021}, this pipeline combines cross-or\-ga\-ni\-za\-tion\-al data to enhance model robustness and scalability, facilitating its deployment in a \ac{wwl} context.

These trajectories capture diverse motion conditions to create a comprehensive dataset. Data is centrally stored, semantically tagged, and reused to avoid redundant collection efforts, aligning with FAIR principles \parencite{Wilkinson2016}. Metadata, including velocity and acceleration scaling factors, robot instance identifiers, and git commit hashes, ensures traceability. Trajectory data is generated for \emph{training and validation}, and  \emph{evaluation} based on tasks proposed by Schneider et al. in~\parencite{Schneider2024}:

\textit{Training and Validation Data}: Planned and executed joint space motions are sampled randomly within the robot's joint limits~\parencite{Schneider2024}, incorporating varied velocity and acceleration profiles to ensure broad workspace coverage.

\textit{Evaluation Data}: Standardized ISO 9283 test paths~\parencite{iso9283} with fixed velocity and acceleration scaling factors of 0.25 provide an industrial scenario for evaluation.

The inverse dynamics problem is reformulated as a data-driven mapping, bypassing the need for explicit computation of dynamic terms such as inertia, Coriolis, and gravity matrices. This approach captures the robot’s dynamics directly from empirical data, simplifying the development process for precise torque-based control.

The \ac{d2k} pipeline for creating the Foundation Model currently operates as follows:
\begin{enumerate}
    \item At 2 a.m. every morning, the current \acsp{ds} for trajectory data are pulled from the data repository, ensuring up-to-date data for training. Statistics of the dataset are analyzed and uploaded back to the data repository. 
    \item A sweep agent is started, which connects to the sweep server. Ten new models are trained, with the agent requesting a new set of hyperparameters from the sweep server for each run. 
    \item If a trained model achieves a lower cross-validation loss than the current model in the model repository, the repository is updated with the new model and its hyperparameters. The updated model is then evaluated on the test data, and the results are stored.
\end{enumerate}

Instance models are fine-tuned on demand by adapting up to five layers of the foundation model, enhancing training efficiency while maintaining high accuracy. This structured approach facilitates rapid deployment and optimization, demonstrating the \ac{d2k} pipeline’s effectiveness in the \acs{wwl} ecosystem.

\subsection{Benchmark}\label{sec:Benchmark}

This benchmark evaluates the performance of four training setups for inverse dynamics modeling: end-to-end training, fine-tuned foundation models, and fine-tuned instance models with known and unknown hyperparameters. The aim is to assess the efficiency and accuracy of fine-tuning approaches compared to traditional end-to-end methods, highlighting the advantages of \ac{d2k} pipelines in collaborative and dynamic industrial environments.

\paragraph{Data Summary}
The benchmark uses a centralized dataset containing 2,533 trajectories and 554,679 measurements aggregated from three institutes: LLT, ITA, and WZL (Table~\ref{tab:summary}). By centralizing and reusing data, the benchmark significantly reduces the time and cost associated with new data collection. This approach underscores the importance of shared repositories in facilitating scalable and efficient knowledge transfer.

\begin{table}[htb]
\centering
\caption{Summary table of the datasets by type and source.}
\begin{tabular}{@{}lrrrr@{}}
\toprule
\textbf{} & \textbf{LLT} & \textbf{ITA} & \textbf{WZL} & \textbf{Total} \\ \midrule
Number of Trajectories & 1,284 & 316 & 933 & 2,533 \\ 
Total Number of Measurements per Axis \hspace{0.5cm} & 230,627 & 87,950 & 236,102 & 554,679 \\ \bottomrule
\end{tabular}

\label{tab:summary}
\end{table}

\begin{figure}[htb]
\centering
\includegraphics[width=\textwidth]{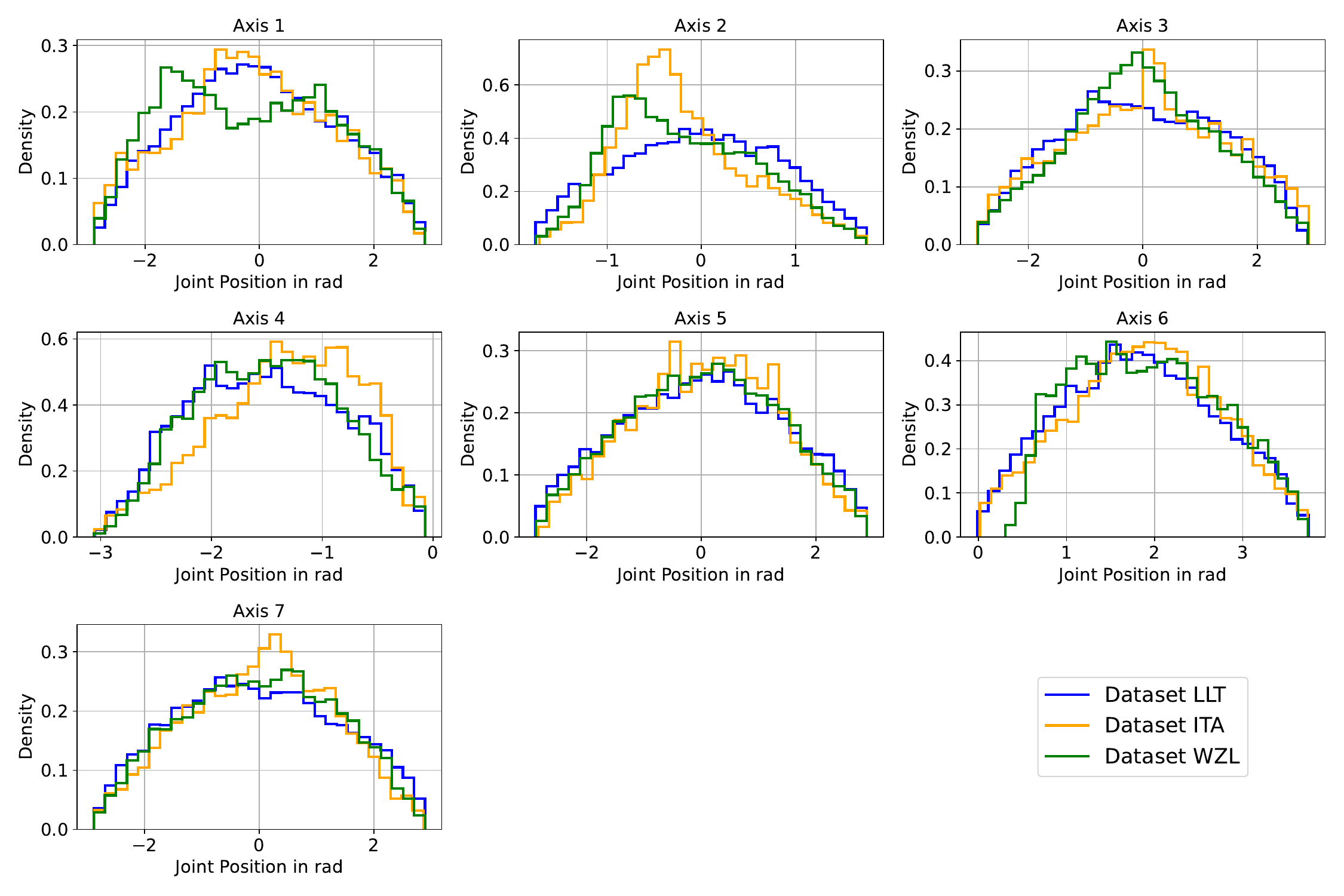}
\caption{Joint position histogram across datasets for each axis. Bars are colourless for readability. The plot for axis two of the robot highlights the effect of workspace restrictions for WZL and ITA on dataset distribution.}
\label{fig:DatasetDistribution}
\end{figure}

The dataset captures diverse operational characteristics, including joint positions, velocities, and accelerations, reflecting variations in workspace constraints across different robot instances. For example, LLT's robot data shows a distribution closer to the target joint-space distribution, as it operates with fewer workspace restrictions. In contrast, ITA and WZL datasets reveal narrower distributions due to specific operational constraints (Figure~\ref{fig:DatasetDistribution}). These variations contribute to the robustness of the aggregated dataset, enabling the foundation model to generalize effectively across diverse conditions.

\begin{figure}[htb]
\centering
\includegraphics[width=\textwidth]{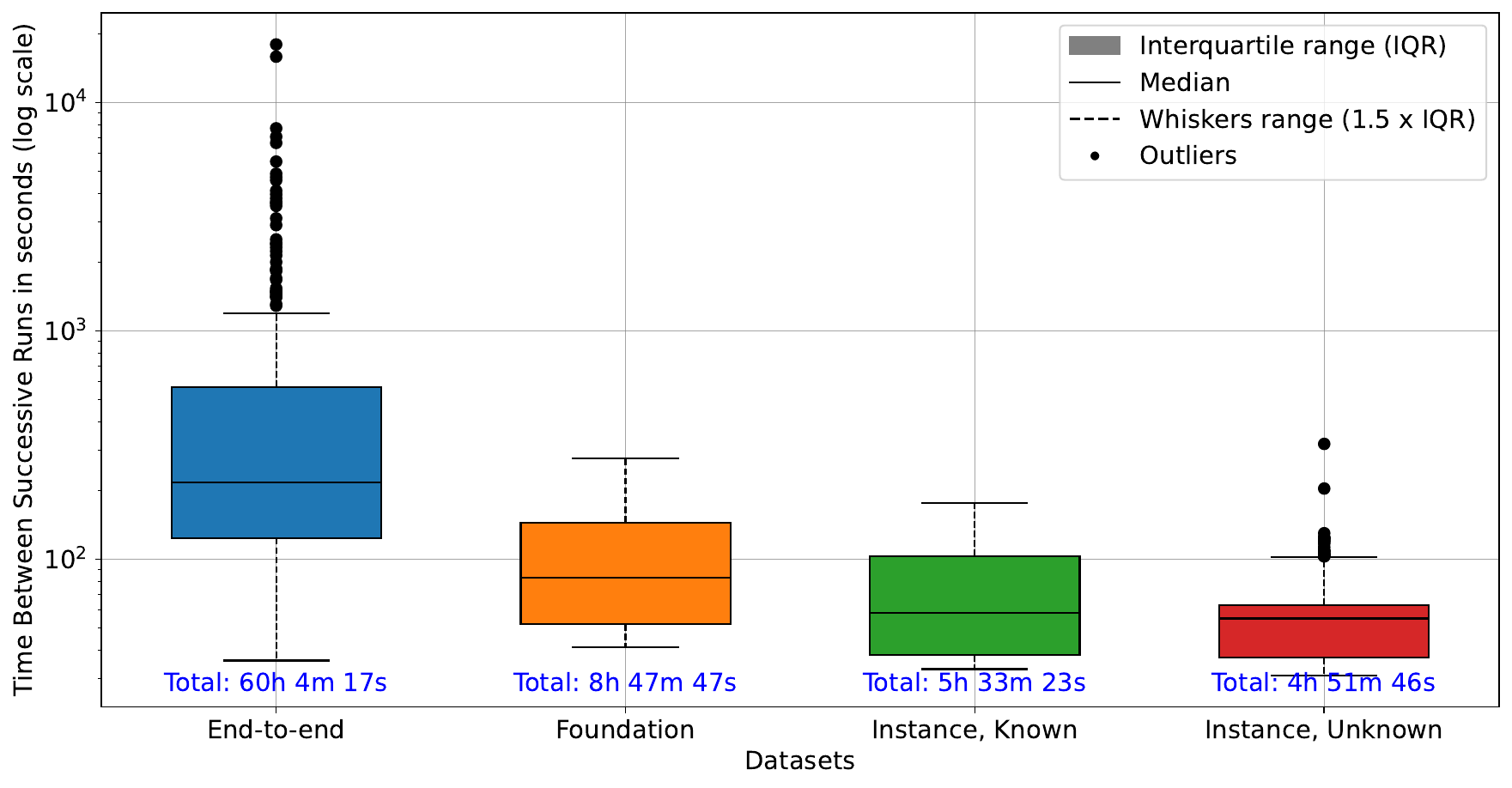}
\caption{Boxplot of runtime distributions for different training setups (log scale). Fine-tuning approaches are faster than end-to-end training in every metric.}
\label{fig:BOX_lots_long}
\end{figure}

\paragraph{Training Time}
Fine-tuned models significantly reduce runtime compared to end-to-end training, showcasing the computational efficiency of \ac{d2k} pipelines. The end-to-end model requires approximately 60 hours to train, with an average runtime of 12 minutes, 3 seconds per run. Fine-tuning the foundation model reduces this time to 8 hours, 47 minutes, averaging 1 minute, 45 seconds per run. Instance-specific fine-tuning with known hyperparameters further decreases runtime to 5 hours, 33 minutes, averaging 1 minute, 6 seconds per run. Fine-tuning an instance model with unknown hyperparameters, completes in just 4 hours, 51 minutes, with an average runtime of 58 seconds per run (Figure~\ref{fig:BOX_lots_long}).

Reusable hyperparameter data enhances this efficiency further, significantly reducing the time required for exhaustive hyperparameter searches, which can take up to 5 hours per run. This reuse aligns with the \ac{wwl}'s goals of ecological sustainability and efficient knowledge transfer, making fine-tuning particularly advantageous for dynamic production environments requiring rapid model adaptation.

\begin{figure}[htb]
\centering
\includegraphics[width=\textwidth]{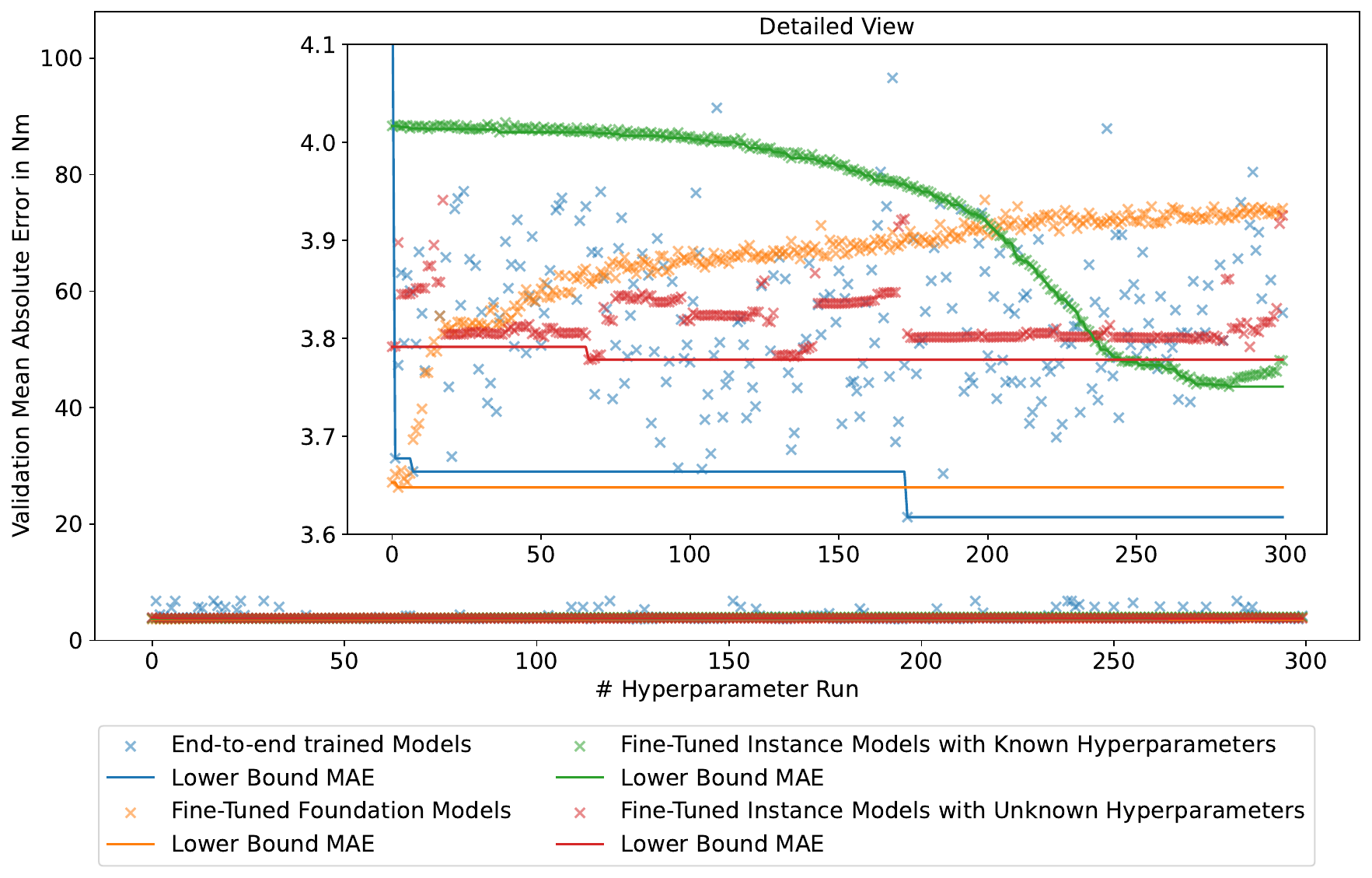}
\caption{Validation Mean Absolute Error (MAE) trends across training setups. The outer plot is scaled to the maximum possible MAE, with an inset detailing the area of interest. Fine-tuning a foundation model (\ac{d2k} approach) performs best initially indicating efficiency. With current hyperparameter refinement implementation, end-to-end training slightly surpasses this MAE after extensive tuning.}
\label{fig:MEA_lots_long}
\end{figure}

\paragraph{Validation Accuracy}
Fine-tuning approaches improve both model convergence and accuracy (compare Figure~\ref{fig:MEA_lots_long}). The validation \ac{mae} is overall close to the theoretically achievable minimum MAE---values below 0.15~Nm are within the torque sensor inaccuracy range---and far from the theoretically achievable maximum MAE of 108.71~Nm. In the detailed view, trends show
that end-to-end training initially has higher MAE values, reflecting slower convergence and outliers indicated bad tries in the hyperparameter search, i.e. inefficient data processing. 
However, with extended training, end-to-end training achieves a similar lower-bound \ac{mae} as the fine-tuned models, demonstrating robust accuracy with sufficient training time.
Fine-tuning the foundation model and instance-specific models with known hyperparameters exhibit faster convergence and lower initial \ac{mae}. Models fine-tuned with unknown hyperparameters display greater variability in early training but eventually achieve comparable accuracy. 

The ability of the foundation model---i.e. our \ac{d2k} approach---to reach $\pm1\%$ of its lower-bound \ac{mae} on the first run without additional hyperparameter refinement highlights the efficiency of \ac{d2k} pipelines, particularly for scenarios requiring rapid deployment and precision.
These results emphasize the advantages of \ac{d2k} pipelines in reducing training time and ensuring high accuracy. By aggregating data and leveraging fine-tuning, these pipelines support scalable and efficient knowledge transfer across diverse industrial applications, validating their potential in modern production environments.

\section{Discussion}\label{section:discussion}

This work introduced \ac{d2k} and \ac{k2d} pipelines for production, utilizing interconnected \acsp{ds} to transform production systems.
By capturing and semantically annotating trajectory data from similar robots across different organizations using \acp{ds}, these pipelines facilitate the training of foundational control models.
We addressed the scarcity of suitable training data not through time-consuming data generation at a single institute, but by leveraging multiple similar machines across different institutes. This approach significantly reduced the time required for data generation and enhanced the quality of the resulting models.
Instead of building new silos for data storage and exchange, we bridged systems by integrating our existing long-term research data management solution and semantic coupling that can easily extended to more use cases.
This demonstrates how innovative digital frameworks can enhance collaboration, optimize production processes, and address challenges in modern manufacturing environments.

The \ac{d2k} and \ac{k2d} pipelines create a seamless, bidirectional flow of information within production environments, transforming raw data into actionable knowledge while allowing existing knowledge to guide data collection and processing.
This synergy promotes scalability and adaptability across production domains, supporting dynamic systems capable of continuous improvement which is mandatory in modern manufacturing systems~\parencite{Koren2017}.
Autonomous agents play a crucial role, ensuring that \ac{d2k} and \ac{k2d} transformations and applications remain adaptive and responsive to changing conditions.
The mutual reinforcement between human expertise and artificial intelligence is essential, as humans provide domain-specific insights while artificial agents enable scalable data processing and predictive capabilities. 
This combination enhances decision-making, improves efficiency, and fosters innovation, providing a strategic advantage to organizations operating in competitive markets.

\acsp{ds} are central to this framework, offering task-specific, context-dependent views of data rather than attempting to comprehensively replicate physical entities , like \acsp{dt} do \parencite{Brauner2022}. This modularity facilitates the integration of diverse systems and abstraction levels, such as individual machine-level \acp{ds} interacting with machine type-level \acp{ds} or task-level \ac{ds}. 
For instance, a \ac{bt} \ac{ds} describing an assembly task workflow can communicate with object 3D model \acsp{ds} to extract grasp poses and forces, while instance-specific dynamics and kinematics \acsp{ds} assist with trajectory generation \parencite{behery2023self}. 
These multi-level \acsp{ds} enhance interoperability and support real-time decision-making, bridging the gap between the physical and digital worlds and ensuring continuous system improvement.

This approach aligns closely with the vision of the \ac{wwl} by enabling cross-organizational collaboration and data sharing. Aggregating diverse production data reduces redundancy, enriches data-driven models, and accelerates technological advancements, as outlined in \parencite{BeheryWWL2023}. However, achieving this collaboration requires balancing data anonymization with utility.
Privacy-preserving techniques \parencite{Sweeney2002, Ninghui2007, Machanavajjhala2006, Zheng2020} and clear governance frameworks \parencite{Abraham2019, Hummel2021} are essential for protecting sensitive information while fostering trust among collaborators~\parencite{Otto2019,Cuñat2024}.
Modular system architectures and adherence to semantic interoperability standards further ease the integration of legacy systems into our proposed pipelines, reducing barriers to entry and encouraging broader participation \parencite{DiMartino2022}.

Increased data integration impacts the workforce.
Well-designed Human-Machine Interfaces (HMIs) minimize automation bias and enhance decision-making \parencite{Endsley2017}, while upskilling employees to interact with these systems ensures successful adoption.
Human-machine collaboration improves safety, efficiency, and job satisfaction, highlighting the importance of human factors in deploying such pipelines and addressing a holistic view of the Internet of Sustainable Production (IoSP) as outlined by Bernhard et al. in~\parencite{Bernhard2023}.

The role of \acsp{ds} in enhancing decision-making processes is critical.
They provide real-time insights and facilitate quick responses to production challenges, supporting operational efficiency and driving strategic innovation \parencite{Pause2019}.
By leveraging knowledge-driven data collection through \ac{k2d} pipelines, organizations can proactively optimize their processes.
Additionally, the pipelines contribute to sustainability by optimizing resource utilization, reducing redundancy, and fostering greener technologies. Their adaptability enhances the resilience of production systems, enabling rapid adjustments to market or environmental changes, aligning with the vision of an IoSP \parencite{Bernhard2023}.

Despite their advantages, these pipelines face challenges, including the need for standardization of data formats and semantic models to ensure interoperability across systems \parencite{Koren2024, TGU2024}.
Data privacy and intellectual property concerns require robust anonymization techniques and trust-building measures \parencite{Brickell2008,Pennekamp2019}.
Scaling the proof-of-concept to include diverse data sources and environments is necessary for broader validation. Incorporating simulations with domain randomization could further extend the approach, supporting more complex and varied production scenarios.

By leveraging \ac{d2k} and \ac{k2d} pipelines built upon \acsp{ds}, organizations can achieve a strategic advantage in innovation, efficiency, and sustainability. This framework supports intelligent, responsive production systems capable of meeting rapidly changing market demands while fostering collaboration and shared value creation in the broader ecosystem of Industry 4.0.

\section{Conclusion and Outlook}\label{section:conclusion}
Our approach aligns with the broader trends of Industry 4.0, complementing ongoing digital transformation efforts.
By implementing \acsp{ds}, \ac{d2k} and \ac{k2d} pipelines, organizations can drive innovation, enhance competitiveness, and explore new business models within these emerging platform-based ecosystems of value co-creation \parencite{Millan2023}.
The framework facilitates the development of intelligent, responsive production systems capable of meeting the demands of rapidly changing markets.

Future research should focus on scaling the proposed approach, including the integration of additional robot instances, other production machinery, or varied production scenarios.
Incorporating continuous learning mechanisms could enhance the pipelines by avoiding automated \enquote{polling} for new data and enabling systems to identify and request the most informative data points.
Benchmarking our approach against other paradigms, such as federated learning, would provide insights into its relative strengths and areas for improvement.
Expanding the pipelines across multiple companies and domains could advance towards more autonomous decision-making processes, further realizing the vision of the \ac{wwl}.

\subsubsection*{Acknowledgements} Funded by the Deutsche Forschungsgemeinschaft (DFG, German Research Foundation) under Germany's Excellence Strategy -- EXC-2023 Internet of Production -- 390621612.

%
%
%
\newpage
\printbibliography
\end{document}